%% file: main.tex
\documentclass[sigconf]{acmart}

\usepackage{makecell}
\usepackage{arydshln}
\usepackage{soul}

\AtBeginDocument{%
  \providecommand\BibTeX{{%
    \normalfont B\kern-0.5em{\scshape i\kern-0.25em b}\kern-0.8em\TeX}}}

\renewcommand\footnotetextcopyrightpermission[1]{} 
\setcopyright{none}

\begin{document}


\title{CharacterFactory: Sampling Consistent Characters with GANs for Diffusion Models}



\author{Qinghe Wang}
\email{qinghewang@mail.dlut.edu.cn}
\affiliation{%
  \institution{Dalian University of Technology}
}

\author{Baolu Li}
\email{8ruceli3@gmail.com}
\affiliation{%
  \institution{Dalian University of Technology}
}

\author{Xiaomin Li}
\email{xmli22@mail.dlut.edu.cn}
\affiliation{%
  \institution{Dalian University of Technology}
}

\author{Bing Cao}
\email{caobing@tju.edu.cn}
\affiliation{%
  \institution{Tianjin University}
}

\author{LiQian Ma}
\email{liqianma.scholar@outlook.com}
\affiliation{%
  \institution{ZMO AI Inc.}
}

\author{Huchuan Lu}
\email{lhchuan@dlut.edu.cn}
\affiliation{%
  \institution{Dalian University of Technology}
}

\author{Xu Jia}
\authornote{Corresponding author.}
\email{jiayushenyang@gmail.com}
\affiliation{%
  \institution{Dalian University of Technology}
}




\begin{abstract}
Recent advances in text-to-image models have opened new frontiers in human-centric generation. However, these models cannot be directly employed to generate images with consistent newly coined identities. In this work, we propose CharacterFactory, a framework that allows sampling new characters with consistent identities in the latent space of GANs for diffusion models. More specifically, we consider the word embeddings of celeb names as ground truths for the identity-consistent generation task and train a GAN model to learn the mapping from a latent space to the celeb embedding space. In addition, we design a context-consistent loss to ensure that the generated identity embeddings can produce identity-consistent images in various contexts. Remarkably, the whole model only takes 10 minutes for training, and can sample infinite characters end-to-end during inference. Extensive experiments demonstrate excellent performance of the proposed CharacterFactory on character creation in terms of identity consistency and editability. Furthermore, the generated characters can be seamlessly combined with the off-the-shelf image/video/3D diffusion models. We believe that the proposed CharacterFactory is an important step for identity-consistent character generation. Project page is available at: \textcolor{magenta}{\url{https://qinghew.github.io/CharacterFactory/}}.
\end{abstract}




\begin{teaserfigure}
\centering
  \includegraphics[width=\textwidth]{"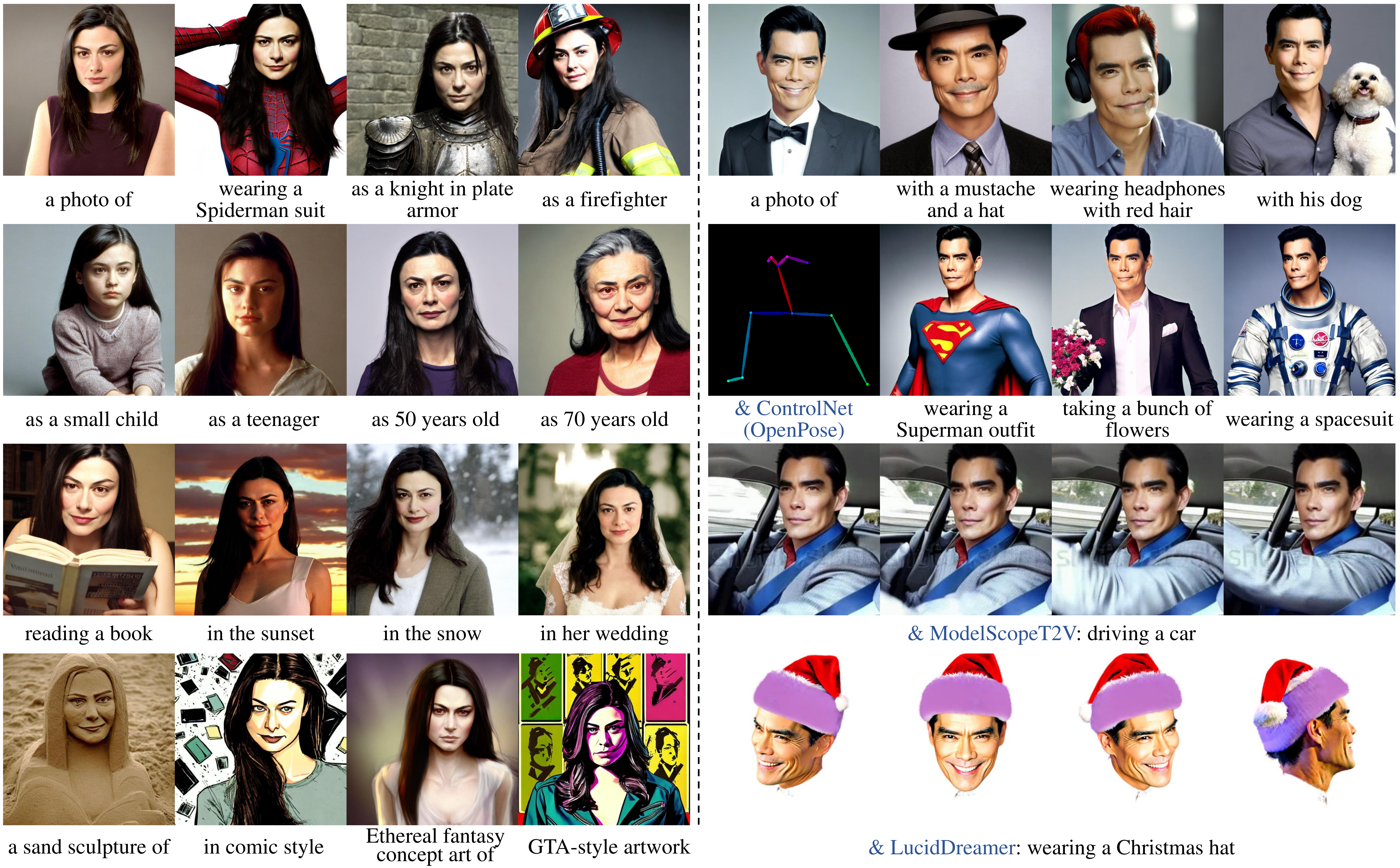"}
  \caption{CharacterFactory can create infinite new characters end-to-end for identity-consistent generation with diverse text prompts. Notably, it can be seamlessly combined with ControlNet~(image), ModelScopeT2V~(video) and LucidDreamer~(3D).}
  \label{fig:teaser}
\end{teaserfigure}



\maketitle

\section{Introduction}
In the evolving realm of text-to-image generation, diffusion models have emerged as indispensable tools for content creation~\cite{rombach2022high,chen2023pixart,zhang2023text}. However, the inherent stochastic nature of the generation models leads to the inability to generate consistent subjects in different contexts directly, as shown in Figure~\ref{fig:teaser}. Such consistency can derive many applications: illustrating books and stories, creating brand ambassador, movie making, developing presentations, art design, identity-consistent data construction and more.

Subject-driven methods work by either representing a user-specific image as a new word~\cite{li2023photomaker,wang2024stableidentity,chen2023dreamidentity} or learning image feature injection~\cite{wang2024instantid,ye2023ip,wei2023elite} for consistent image generation. Their training paradigms typically include per-subject optimization and encoder pretraining on large-scale datasets. The former usually requires lengthy optimization for each subject and tends to overfit the appearance in the input image~\cite{gal2022image,ruiz2023dreambooth}. 
The latter consumes significant computational costs and struggles in stably capturing the identity and its details~\cite{li2023photomaker,wang2024instantid}. However, these methods attempt to produce images with the same identity as the reference images, instead of creating a new character in various contexts. A feasible way is that a text-to-image model is used in advance to create a new character's image and then subject-driven methods are adopted to produce images with consistent identity. Such a two-stage workflow could push the pretrained generation model away from its training distribution, leading to degraded generation quality and poor compatibility with other extension models. Therefore, there is a pressing need to propose a new end-to-end framework that enables consistent character generation.

Here we are particularly interested in consistent image generation for human. Since text-to-image models are pretrained on large-scale image-text data, which contains massive text prompts with celeb names, the models can generate identity-consistent images using celeb names. These names are ideal examples for this task. Previous work~\cite{wang2024stableidentity} has revealed that the word embeddings of celeb names constitute a human-centric prior space with editability, so we decide to conduct new character sampling in this space.

In this work, we propose CharacterFactory, a framework for new character creation which mainly consists of an Identity-Embedding GAN~(IDE-GAN) and a context-consistent loss. Specifically, a GAN model composed of MLPs is used to map from a latent space to the celeb embedding space following the adversarial learning manner, with word embeddings of celeb names as real data and generated ones as fake. Furthermore, to enable the generated embeddings to work like the native word embeddings of CLIP~\cite{radford2021learning}, we constrain these embeddings to exhibit consistency when combined with diverse contexts. Following this paradigm, the generated embeddings could be naturally inserted into CLIP text encoder, hence could be seamlessly integrated with the image/video/3D diffusion models. In addition, since IDE-GAN is composed of only MLPs as trainable parameters and accesses only the pretrained CLIP during training, it takes only 10 minutes to train and then infinite new identity embeddings could be sampled to produce identity-consistent images for new characters during inference.

The main contributions of this work are summarized as follows: 1) We for the first time propose an end-to-end identity-consistent generation framework named CharacterFactory, which is empowered by a vector-wise GAN model in CLIP embedding space. 2) We design a context-consistent loss to ensure that the generated pseudo identity embeddings can manifest contextual consistency. This plug-and-play regularization can contribute to other related tasks. 3) Extensive experiments demonstrate superior identity consistency and editability of our method. In addition, we show the satisfactory interpolation property and strong generalization ability with the off-the-shelf image/video/3D modules.

\section{Related Work}
\subsection{Text-to-Image Diffusion Models}
Recent advances in diffusion models~\cite{ho2020denoising,song2020denoising} have shown unprecedented capabilities for text-to-image generation~\cite{rombach2022high,nichol2021glide,ramesh2022hierarchical}, and new possibilities are still emerging~\cite{wang2022hs,chen2024shoemodel}. The amazing generation performance is derived from the high-quality large-scale image-text pairs~\cite{schuhmann2022laion,schuhmann2021laion}, flourishing foundational models~\cite{chen2023pixart,peebles2023scalable}, and stronger controllability design~\cite{zhang2023adding,zhang2024transparent}. Their fundamental principles are based on Denoising Diffusion Probabilistic Models~(DDPMs)~\cite{ho2020denoising}, which include a forward noising process and a reverse denoising process. The forward process adds Gaussian noise progressively to an input image, and the reverse process is modeled with a UNet trained for predicting noise. Supervised by the denoising loss, a random noise can be denoised to a realistic image by iterating the reverse diffusion process. However, due to the stochastic nature of this generation process, existing text-to-image diffusion models are not able to directly implement consistent character generation.

\begin{figure*}[!t]
  \centering
  \includegraphics[width=1\linewidth]{"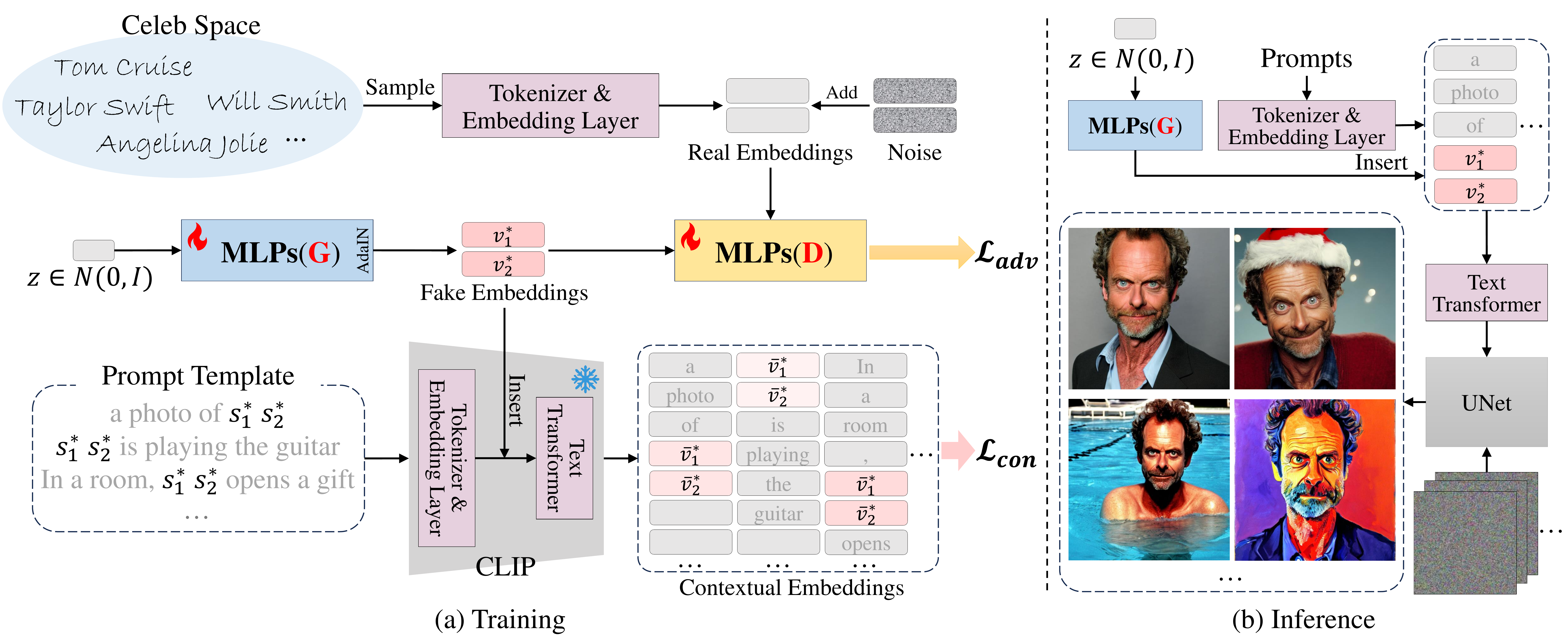"}
      \caption{Overview of the proposed \textit{CharacterFactory}. (a) We take the word embeddings of celeb names as ground truths for identity-consistent generation and train a GAN model constructed by MLPs to learn the mapping from $z$ to celeb embedding space. In addition, a context-consistent loss is designed to ensure that the generated pseudo identity can exhibit consistency in various contexts. $s_1^*$, $s_2^*$ are placeholders for $v_1^*$, $v_2^*$. (b) Without diffusion models involved in training, IDE-GAN can end-to-end generate embeddings that can be seamlessly inserted into diffusion models to achieve identity-consistent generation.}
  \label{fig:framework}
\end{figure*}

\subsection{Consistent Character Generation}
\label{sec:related_consistent}
Existing works on consistent character generation mainly focus on personalization for the target subject~\cite{gal2022image,ruiz2023dreambooth}. 
Textual Inversion~\cite{gal2022image} represents the target subject as a new word embedding via optimization while freezing the diffusion model. DreamBooth~\cite{ruiz2023dreambooth} finetunes all weights of the diffusion model to fit only the target subject. IP-Adapter~\cite{ye2023ip} designs a decoupled cross-attention mechanism for text features and image features. Celeb-Basis~\cite{yuan2023inserting} and StableIdentity~\cite{wang2024stableidentity} use prior information from celeb names to make optimization easier and improve editability. PhotoMaker trains MLPs and the LoRA residuals of the attention layers to inject identity information~\cite{li2023photomaker}. But these methods attempt to produce identity-consistent images based on the reference images, instead of creating a new character. In addition, The Chosen One~\cite{avrahami2023chosen} clusters the generated images to obtain similar outputs for learning a customized model on a highly similar cluster by iterative optimization with personalized LoRA weights and word embeddings, which is a time-consuming process. ConsiStory~\cite{tewel2024training} introduces a shared attention block mechanism and correspondence-based feature injection between a batch of images, but relying only on patch features lacks semantic understanding for the subject and makes the inference process complicated. Despite creating new characters, they still suffer from complicated pipelines and poor editability.

\subsection{Integrating Diffusion Models and GANs}
Generative Adversarial Net (GAN)~\cite{goodfellow2014generative, karras2019style} models the mapping between data distributions by adversarially training a generator and a discriminator. Although GAN-based methods have been outperformed by powerful diffusion models for image generation, they perform well on small-scale datasets~\cite{dhariwal2021diffusion} benefiting from the flexibility of GANs. Some methods focus on combining them to improve the optimization objective for diffusion models with GANs~\cite{xu2023ufogen,xiao2021tackling,wang2022diffusion}. In this work, we for the first time construct a GAN model in CLIP embedding space to sample consistent identity for diffusion models.

\section{Method}
To enable the text-to-image models to directly generate images with the same identity, we present a new end-to-end framework, named \textit{CharacterFactory}, which produces pseudo identity embeddings that can be inserted into any contexts to achieve identity-consistent character generation, as shown in Figure~\ref{fig:framework}. In this section, the background of Stable Diffusion is first briefly introduced in Section~\ref{sec:3.1}. Later, the technical details of the proposed CharacterFactory are elaborated in Section~\ref{sec:3.2} and~\ref{sec:3.3}. Finally, our full objective is demonstrated in Section~\ref{sec:3.4}.

\subsection{Preliminary}
\label{sec:3.1}
In this work, we employ the pretrained Stable Diffusion~\cite{rombach2022high}~(denoted as SD) as the base text-to-image model. SD consists of three components: a CLIP text encoder $e_{text}$~\cite{radford2021learning}, a Variational Autoencoder~(VAE)~($\mathcal{E}$, $\mathcal{D}$)~\cite{esser2021taming} and a denoising U-Net $\epsilon_\theta$. With the text conditioning, $\epsilon_\theta$ can denoise sampled Gaussian noises to realistic images conforming to the given text prompts $p$. 
In particular, the tokenizer of $e_{text}$ sequentially divides and encodes $p$ into $l$ integer tokens. Subsequently, by looking up the tokenizer's dictionary, the embedding layer of $e_{text}$ retrieves a group of corresponding word embeddings $g = [v_1, ..., v_l], v_i\in\mathbb{R}^d$. 
Then, the text transformer $\tau_{text}$ of $e_{text}$ further represents $g$ to contextual embeddings $\bar{g} = [\bar{v}_1, ..., \bar{v}_l], \bar{v}_i\in\mathbb{R}^d$ with the cross-attention mechanism. And $\epsilon_\theta$ renders the content conveyed in text prompts by cross attention between $\bar{g}$ and diffusion features.

\subsection{IDE-GAN}
\label{sec:3.2}
Since Stable Diffusion is trained with numerous celeb photos and corresponding captions with celeb names, these names can be inserted into various contexts to generate identity-aligned images. We believe that the word embeddings of these celeb names can be considered as ground truths for identity-consistent generation. Therefore, we train an Identity-Embedding GAN~(IDE-GAN) model to learn a mapping from a latent space to the celeb embedding space, $G: z\rightarrow v$, with the expectation that it can generate pseudo identity embeddings that master the identity-consistent editability, like celeb embeddings.

\begin{figure}[!t]
  \centering
  \includegraphics[width=1\linewidth]{"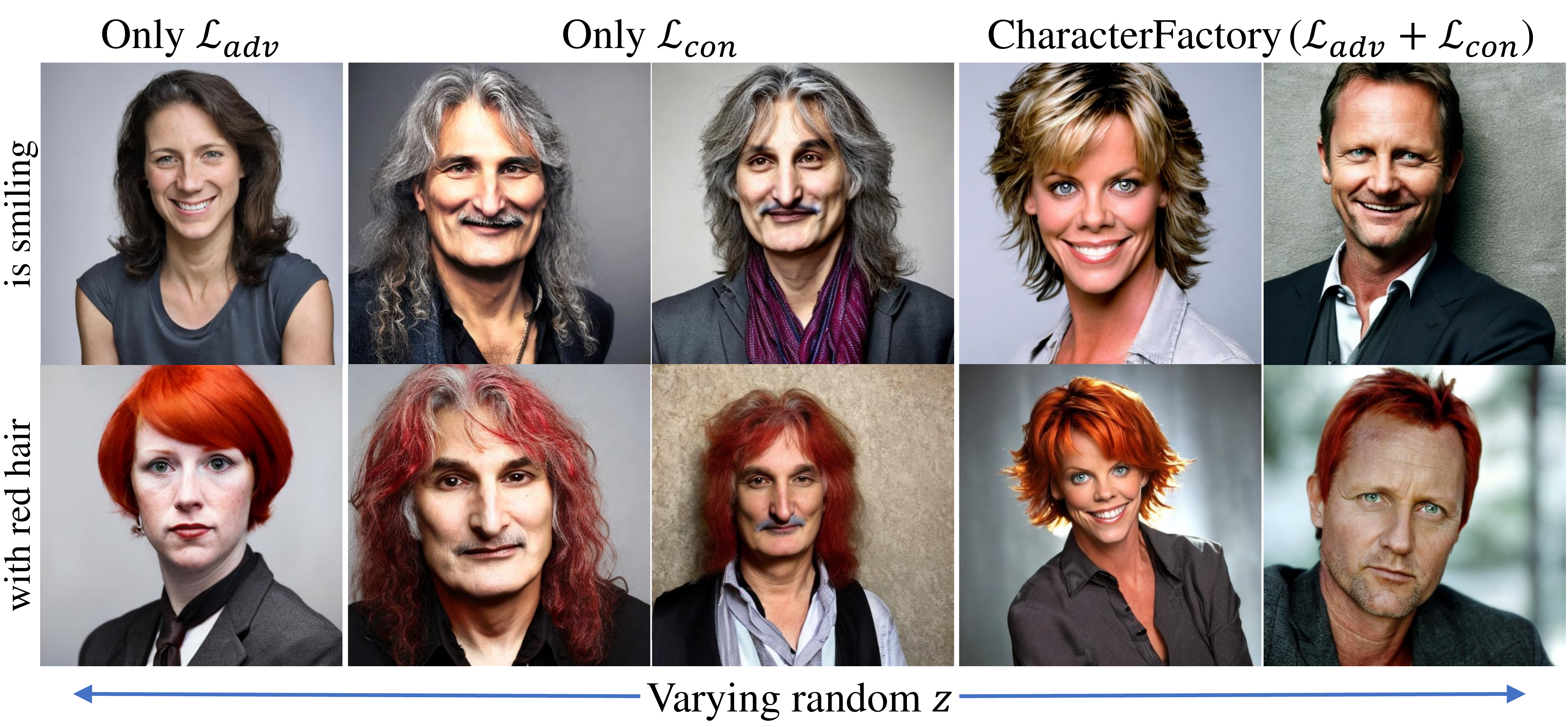"}
    \caption{Effect of $\mathcal{L}_{adv}$ and $\mathcal{L}_{con}$. The images in each column are generated by a randomly sampled $z$ and two prompts according to the pipeline in Figure~\ref{fig:framework}(b). The placeholders $s_1^*$, $s_2^*$ of prompts such as ``$s_1^*$ $s_2^*$ is smiling'' are omitted in this work for brevity~(\textit{Zoom in for the best view}).}
  \label{fig:component_effect}
\end{figure}

Specifically, we employ 326 celeb names~\cite{wang2024stableidentity} which consist only of first name and last name, and encode them into the corresponding word embeddings $C\in\mathbb{R}^{326\times 2\times d}$ for training. In addition, we observe that adding a small noise to the celeb embeddings can still generate images with corresponding identity. Therefore, we empirically introduce random noise $\eta\sim\mathcal{N}(0,\mathbf{I})$ scaled by $5e-3$ as a data augmentation. As shown in Figure~\ref{fig:framework}(a), given a latent code $z\in\mathcal{N}(0,\mathbf{I})$, the generator $G$ is trained to produce embeddings $[v_1^*, v_2^*]$ that cannot be distinguished from ``real''~(i.e., celeb embeddings) by an adversarially trained discriminator $D$. To alleviate the training difficulty of $G$, we use AdaIN to help the MLPs' output embeddings $[v'_1, v'_2]$ land more naturally into the celeb embedding space~\cite{wang2024stableidentity}:
\begin{equation}
\label{adain}
    v^*_i = \sigma(C_i)(\frac{v'_i - \mu(v'_i)}{\sigma(v'_i)}) + \mu(C_i), for~i = 1,2
\end{equation}
where $\mu(v'_i), \sigma(v'_i)$ are scalars. $\mu(C_i)\in\mathbb{R}^d$, $\sigma(C_i)\in\mathbb{R}^d$ are vectors, because each dimension of $C_i$ has a different distribution. And $D$ is trained to detect the generated embeddings as ``fake''. This adversarial training is supervised by: 
\begin{equation}
\label{L_gan}
     \mathcal{L}_{adv} = \mathbb{E}_{[v_1, v_2]\sim C}[\log D([v_1, v_2] + \eta)] + \mathbb{E}[\log(1-D(G(z)))],
\end{equation}
where $G$ tries to minimize this objective and $D$ tries to maximize it. As shown in the column 1 of Figure~\ref{fig:component_effect}, $[v_1^*, v_2^*]$ generated by $z$ can be inserted into different contextual prompts to produce human images while conforming to the given text descriptions. It indicates that $[v_1^*, v_2^*]$ have obtained editability and enough information for human character generation, and flexibility to work with other words for editing, but the setting of ``Only $\mathcal{L}_{adv}$'' can not guarantee identity consistency in various contexts.

\subsection{Context-Consistent Loss}
\label{sec:3.3}
To enable the generated embeddings $[v_1^*, v_2^*]$ to be naturally inserted into the pretrained Stable Diffusion, they are encouraged to work as similarly as possible to normal word embeddings. CLIP, which is trained to align images and texts, could map the word corresponding to a certain subject in various contexts to similar representations. Hence, we design the context-consistent loss to encourage the generated word embeddings to own the same property.

Specifically, we sample 1,000 text prompts with ChatGPT~\cite{ouyang2022training} for various contexts~(covering expressions, decorations, actions, attributes, and backgrounds), like ``Under the tree, $s_1^*$ $s_2^*$ has a picnic'', and demand that the position of ``$s_1^*$ $s_2^*$'' in the context should be as diverse as possible. During training, we sample $N$ prompts from the collected prompt set, and use the tokenizer and embedding layer to encode them into $N$ groups of word embeddings. The generated embeddings $[v_1^*, v_2^*]$ are inserted at the position of ``$s_1^*$ $s_2^*$''. Then, the text transformer $\tau_{text}$ further represents them to $N$ groups of contextual embeddings, where we expect to minimize the average pairwise distances among the $\{[\bar{v}_1^*, \bar{v}_2^*]_i\}_{i=1}^N$:
\begin{equation}
\mathcal{L}_{con} = \frac{1}{\binom{N}{2}}\sum_{j=1}^{N-1} \sum_{k=j+1}^{N} \|[\bar{v}_1^*, \bar{v}_2^*]_j - [\bar{v}_1^*, \bar{v}_2^*]_k\|^2_2 ,
\end{equation}
where $N$ is $8$ as default. In this way, the pseudo word embeddings $[v_1^*, v_2^*]$ generated by IDE-GAN can exhibit consistency in various contexts. A naive idea is to train MLPs with only $\mathcal{L}_{con}$, which shows promising consistency as shown in the column 2,~3 of Figure~\ref{fig:component_effect}. However, $\mathcal{L}_{con}$ only focuses on consistency instead of diversity, mode collapse occurs in spite of different $z$. When $\mathcal{L}_{con}$ and $\mathcal{L}_{adv}$ work together, the proposed CharacterFactory can sample diverse context-consistent identities as shown in the column 4,~5 of Figure~\ref{fig:component_effect}. Notably, this regularization loss is plug-and-play and can contribute to other subject-driven generation methods to learn context-consistent subject word embeddings.

\subsection{Full Objective}
\label{sec:3.4}
Our full objective can be expressed as:
\begin{align}
G^* = \arg\min_{G} \max_{D} \lambda_1\mathcal{L}_{adv}(G, D) + \lambda_2\mathcal{L}_{con}(G, \tau_{text}),
\end{align}
where $\lambda_1$ and $\lambda_2$ are trade-off parameters. The discriminator $D$'s job remains unchanged, and the generator $G$ is tasked not only to learn the properties of celeb embeddings to deceive the $D$, but also to manifest contextual consistency in the output space of the text transformer $\tau_{text}$. Here, we emphasize two noteworthy points:
\begin{itemize}
\item \textbf{GAN for word embedding.} We introduce GAN in the CLIP embedding space for the first time and leverage the subsequent network to design the context-consistent loss which can perceive the generated pseudo identity embeddings in diverse contexts. This design is similar to the thought of previous works for image generation~\cite{isola2017image,zhang2018unreasonable,cao2023multi}, which have demonstrated that mixing the GAN objective and a more traditional loss such as L2 distance is beneficial.

\item \textbf{No need diffusion-based training.} Obviously, the denoising UNet and the diffusion loss which are commonly used to train diffusion-based methods, are not involved in our training process. Remarkably, the proposed IDE-GAN can seamlessly integrate with diffusion models to achieve identity-consistent generation for inference as shown in Figure~\ref{fig:framework}(b).

\end{itemize}

\begin{figure*}[!t]
  \centering
  \includegraphics[width=1\linewidth]{"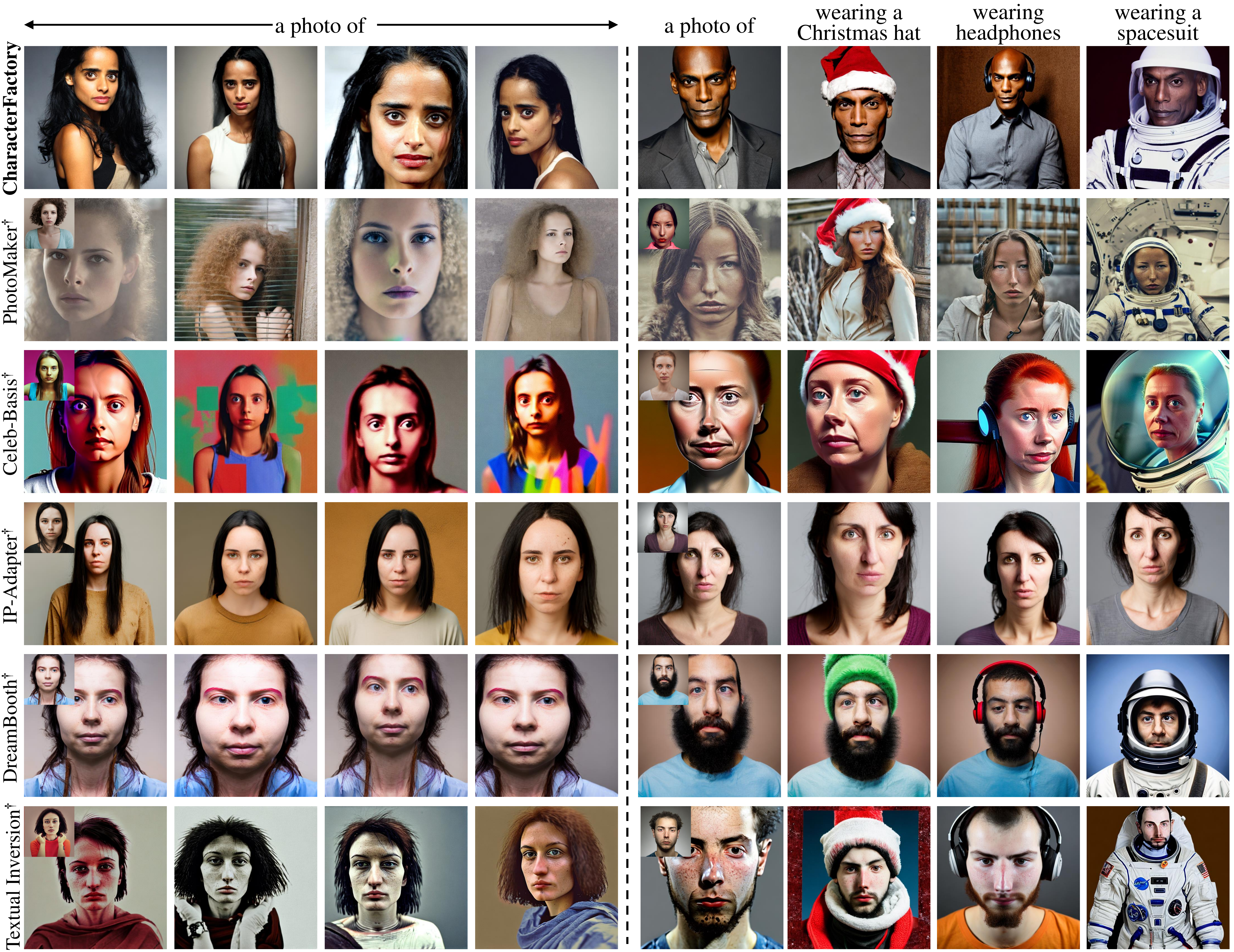"}
    \caption{Qualitative comparisons with two-stage workflows using five baselines (denoted with $\dagger$) for creating consistent characters. The upper left corner of the two-stage baselines is the generated image by Stable Diffusion as the input of the second stage. Two-stage workflows struggle to maintain the identity of the generated image and degrade the image quality. In comparison, the proposed CharacterFactory can generate high-quality identity-consistent character images with diverse layouts while conforming to the given text prompts~(\textit{Zoom in for the best view}).}
  \label{fig:two_stage_comparison}
\end{figure*}

\section{Experiments}
\subsection{Experimental Setting}
\label{sec:4.1}
\noindent\textbf{Implementation Details.}
We employ Stable Diffusion v2.1-base as our base model. The number of layers in the MLPs for the generator $G$ and the discriminator $D$ are 2 and 3 respectively. The dimension of $z$ is set to 64 empirically. The batch size and learning rate are set to 1 and $5e-5$. We employ an Adam optimizer~\cite{kingma2014adam} with the momentum parameters $\beta_1= 0.5$ and $\beta_2 = 0.999$ to optimize our IDE-GAN. The trade-off parameters $\lambda_1$ and $\lambda_2$ are both 1 as default. CharacterFactory is trained with only 10 minutes for 10,000 steps on a single NVIDIA A100. The classifier-free guidance~\cite{ho2022classifier} scale is 8.5 for inference as default. More implementation details can be found in the supplementary material.

\begin{table*}[t]
\caption{Quantitative comparisons with two-stage workflows using five baselines~(denoted with $\dagger$). $\uparrow$ indicates higher is better, and $\downarrow$ indicates that lower is better. The best results are shown in \underline{\textbf{bold}}. We define the speed as the time it takes to create a new consistent character on a single NVIDIA A100 GPU. Obviously, CharacterFactory obtains superior performance on identity consistency, editability, trusted face diversity, image quality and speed, which are consistent with the qualitative comparisons.
}
\label{tab:two_stage_comparison}
\begin{tabular}{lccccccc}
\toprule
                         Methods & Subject Cons.$\uparrow$ & Identity Cons.$\uparrow$ & Editability$\uparrow$    & Face Div.$\uparrow$ & Trusted Div.$\uparrow$ & Image Quality$\downarrow$  & Speed (s)$\downarrow$  \\
\midrule     
Textual Inversion$^\dagger$~\cite{gal2022image}        & 0.647               & 0.295                & 0.274          & \underline{\textbf{0.392}} & 0.078             & 47.94          & 3200       \\
DreamBooth$^\dagger$~\cite{ruiz2023dreambooth}               & 0.681               & 0.443                & 0.287          & 0.339          & 0.073             & 62.66          & 1500       \\
IP-Adapter$^\dagger$~\cite{ye2023ip}               & \underline{\textbf{0.853}}      & 0.447                & 0.227          & 0.192          & 0.096             & 95.25          & 7          \\
Celeb-Basis$^\dagger$~\cite{yuan2023inserting}              & 0.667               & 0.369                & 0.273          & 0.378          & 0.101             & 56.43          & 480        \\
PhotoMaker$^\dagger$~\cite{li2023photomaker}               & 0.694               & 0.451                & 0.301          & 0.331          & 0.138             & 53.37          & 10         \\
\textbf{CharacterFactory} & 0.764               & \underline{\textbf{0.498}}       & \underline{\textbf{0.332}} & 0.333          & \underline{\textbf{0.140}}    & \underline{\textbf{22.58}} & \underline{\textbf{3}}\\
\bottomrule
\end{tabular}
\end{table*}

\begin{figure*}[!t]
  \centering
  \includegraphics[width=0.94\linewidth]{"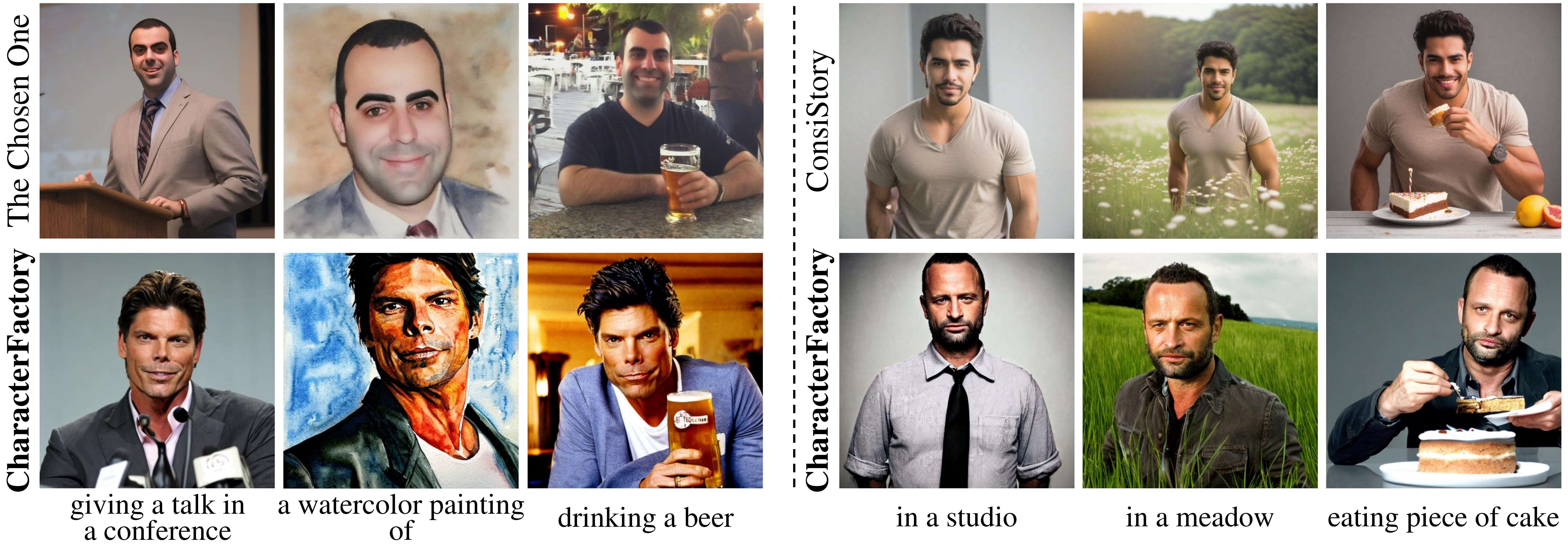"}
    \caption{Qualitative comparisons with the generation results in the papers of two most related methods The Chosen One~\cite{avrahami2023chosen} and ConsiStory~\cite{tewel2024training}. CharacterFactory achieves comparable performance with the same prompts~(\textit{Zoom in for the best view}).}
  \label{fig:chosen_one_consistory}
\end{figure*}

\noindent\textbf{Baselines.} Since the most related methods, The Chosen One~\cite{avrahami2023chosen} and ConsiStory~\cite{tewel2024training} which are also 
designed for consistent text-to-image generation, have not released their codes yet, we compare these methods with the content provided in their papers. In addition, as we introduced in Section 1, the two-stage workflows with subject-driven methods can also create new characters. Therefore, we first use a prompt ``a photo of a person, facing to the camera'' to drive Stable Diffusion to generate images of new characters as the input of the second stage, and then use these subject-driven methods to produce character images with diverse prompts for comparison. \textit{These input images are used for subject information injection and not involved in the calculation of quantitative comparisons.} These methods include the optimization-based methods: Textual Inversion~\cite{gal2022image}, DreamBooth~\cite{ruiz2023dreambooth}, Celeb-Basis~\cite{yuan2023inserting}, and the encoder-based methods: IP-Adapter~\cite{ye2023ip}, PhotoMaker~\cite{li2023photomaker}. We prioritize to use the official models released by these methods. We use the Stable Diffusion 2.1 versions of Textual Inversion and DreamBooth for fair comparison.

\noindent\textbf{Evaluation.} The input of our method comes from random noise, so this work does not compare subject preservation for quantitative comparison. To conduct a comprehensive evaluation, we use 40 text prompts that cover decorations, actions, expressions, attributes and backgrounds~\cite{li2023photomaker}. Overall, we use 70 identities and 40 text prompts to generate 2,800 images for each competing method.

Metrics: We calculate the CLIP visual similarity (CLIP-I) between the generated results of ``a photo of $s_1^*$ $s_2^*$'' and other text prompts to evaluate \textbf{Subject Consistency}. And we calculate face similarity~\cite{deng2019arcface} and perceptual similarity (i.e., LPIPS)~\cite{zhang2018perceptual} between the detected face regions with the same settings to measure the \textbf{Identity Consistency} and \textbf{Face Diversity}~\cite{li2023photomaker,wu2023singleinsert}. But inconsistent faces might obtain high face diversity, leading to unreliable results. Therefore, we also introduce the \textbf{Trusted Face Diversity}~\cite{wang2024stableidentity} which is calculated by the product of cosine distances from face similarity and face diversity between each pair of images, to evaluate whether the generated faces from the same identity are both consistent and diverse. We calculate the text-image similarity~(CLIP-T) to measure the \textbf{Editablity}. In addition, we randomly sample 70 celeb names to generate images with the introduced 40 text prompts as pseudo ground truths, and calculate Fréchet Inception Distance (FID)~\cite{lucic2017gans} between the generated images by competing methods and pseudo ground truths to measure the \textbf{Image Quality}.

\subsection{Comparison with Two-Stage Workflows.}
\noindent\textbf{Qualitative Comparison.}
As mentioned in Section~\ref{sec:4.1}, we randomly generate 70 character images in front view to inject identity information for two-stage workflows using subject-driven baselines~(denoted with $\dagger$), as shown in Figure~\ref{fig:two_stage_comparison}. PhotoMaker$^\dagger$~\cite{li2023photomaker} and Celeb-Basis$^\dagger$~\cite{yuan2023inserting} are human-centric methods. The former pretrains a face encoder and LoRA residuals on large-scale datasets. The latter optimizes word embeddings to represent the target identity. But they all suffer from degraded image quality under this setting. IP-Adapter$^\dagger$~\cite{ye2023ip} learns text-image decoupled cross attention, but fails to present ``Christmas hat'' and ``spacesuit''. DreamBooth$^\dagger$~\cite{ruiz2023dreambooth} finetunes the whole model to adapt to the input image and tends to generate images similar to the input image. It lacks generation diversity and fails to produce the ``Christmas hat''. Due to the stochasticity of Textual Inversion$^\dagger$~\cite{gal2022image}'s optimization process, its identity consistency and image quality are relatively weak. Overall, two-stage workflows show decent performance for identity consistency, editability, and image quality, and they all rely on the input images and struggle to preserve the input identity. In contrast, the proposed CharacterFactory can sample pseudo identities end-to-end and generate identity-consistent prompt-aligned results with high quality.

\begin{figure*}[!t]
  \centering
  \includegraphics[width=1\linewidth]{"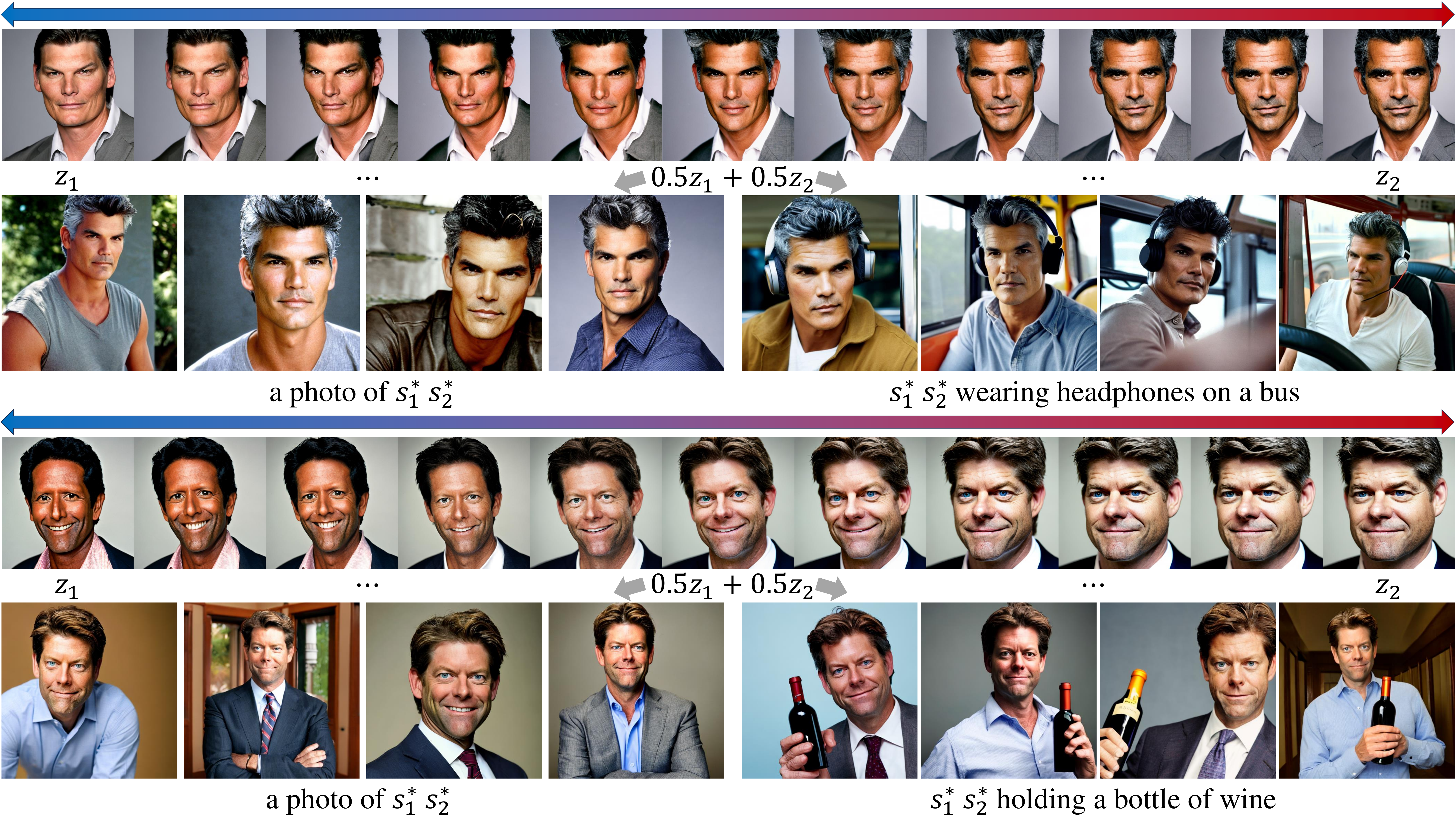"}
    \caption{Interpolation property of IDE-GAN. We conduct linear interpolation between randomly sampled $z_1$ and $z_2$, and generate pseudo identity embeddings with IDE-GAN. To visualize the smooth variations in image space, we insert the generated embeddings into Stable Diffusion via the pipeline of Figure~\ref{fig:framework}(b). The experiments in row 1, 3 are conducted with the same seeds, and row 2, 4 use random seeds~(\textit{Zoom in for the best view}).}
  \label{fig:interpolation}
\end{figure*}

\begin{table}[t]
\caption{Comparisons with two most related methods on the speed~(i.e., time to produce consistent identity) and the forms of identity representation. In contrast, CharacterFactory is faster, and uses a more lightweight and natural form for identity representation, which ensures seamless collaboration with other modules and convenient identity reuse.}
\label{tab:speed_forms}
\setlength{\tabcolsep}{1.7pt}
\begin{tabular}{lcc}
\toprule
                         & Speed$\downarrow$~(s)            & Identity Representation                            \\
\midrule                         
The Chosen One~\cite{avrahami2023chosen}           & 1,200        & LoRAs + two word embeddings                     \\
\hdashline\specialrule{0em}{1pt}{1pt}
Consistory~\cite{tewel2024training} & 49 & \makecell[c]{Self-attention keys and values\\ of reference images} \\
\hdashline\specialrule{0em}{1pt}{1pt}
\textbf{CharacterFactory} & \textbf{3} & \textbf{Two word embeddings}\\
\bottomrule
\end{tabular}
\end{table}

\noindent\textbf{Quantitative Comparison.} In addition, we also provide the quantitative comparison with five baselines in Table~\ref{tab:two_stage_comparison}. Since IP-Adapter$^\dagger$ tends to generate frontal faces, it obtains better subject consistency~(CLIP-I) but weak editability~(CLIP-T). CLIP-I mainly measures high-level semantic alignment and lacks the assessment for identity, so we further introduce the identity consistency for evaluation. Our method achieves the best identity consistency, editability and second-place subject consistency. In particular, the proposed context-consistent loss incentivizes pseudo identities to exhibit consistency in various contexts. On the other hand, our effective adversarial learning enables pseudo identity embeddings to work in Stable Diffusion as naturally as celeb embeddings, and thus outperforms PhotoMaker$^\dagger$~(the second place) by 0.031 on editability. Textual Inversion$^\dagger$ and Celeb-Basis$^\dagger$ obtain good face diversity but weak trusted diversity. This is because face diversity measures whether the generated faces from the same identity are diverse in different contexts, but inconsistent identities can also be incorrectly recognized as ``diverse''. Therefore, trusted face diversity is introduced to evaluate whether the results are both consistent and diverse. So Textual Inversion$^\dagger$ obtains the best face diversity, but is inferior to CharacterFactory 0.062 on trusted face diversity. For image quality~(FID), the two-stage workflows directly lead to an unacceptable degradation of competing methods on image quality quantitatively. On the other hand, two-stage workflows consume more time for creating identity-consistent characters. In comparison, our end-to-end framework implements more natural generation results, the best image quality and faster inference workflow.

\subsection{Comparison with Consistent-T2I Methods}
In addition, we compare the most related methods The Chosen One~\cite{avrahami2023chosen} and ConsiStory~\cite{tewel2024training} with the content provided in their papers. These two methods are also designed for consistent character generation, but have not released the codes yet.

\noindent\textbf{Qualitative Comparison.} As shown in Figure~\ref{fig:chosen_one_consistory}, The Chosen One uses Textual Inversion$+$DreamBooth-LoRA to fit the target identity, but only achieves consistent face attributes, which fails to obtain better identity consistency. Besides, excessive additional parameters degrade the image quality. ConsiStory elicits consistency by using shared attention blocks to learn the subject patch features within a batch. Despite its consistent results, it lacks controllability and semantic understanding of the input subject due to its dependence on patch features, i.e., it cannot edit with abstract attributes such as age and fat/thin. In comparison, our method achieves comparable performance on identity consistency, and image quality, and even can prompt with abstract attributes as shown in Figure~\ref{fig:teaser},~\ref{fig:story}.

\noindent\textbf{Practicality.} As introduced in Section~\ref{sec:related_consistent}, The Chosen One searches a consistent character by a lengthy iterative procedure which takes about 1,200 seconds on a single NVIDIA A100 GPU, and needs to save LoRA weights$+$two word embeddings for each character. ConsiStory is training-free, but its inference pipeline is time-consuming~(takes about 49 seconds to produce an identity-consistent character) and requires saving self-attention keys and values of reference images for each character. In comparison, CharacterFactory is faster and more lightweight, taking only 10 minutes to train IDE-GAN for sampling pseudo identity embeddings infinitely, and only takes 3 seconds to create a new character with Stable Diffusion. Besides, using two word embeddings to represent consistent identity is convenient for identity reuse and integration with other modules such as video/3D generation models.

\subsection{Ablation Study}
In addition to the ablation results presented in Figure~\ref{fig:component_effect}, we also conduct a more comprehensive quantitative analysis in Table~\ref{tab:ablation}. To evaluate the diversity of generated identities, we calculate the average pairwise face similarity between 70 generated images with ``a photo of $s_1^*$ $s_2^*$'', and define $(1 -$ the average similarity$)$ as identity diversity~(The lower similarity between generated identities represents higher diversity). Note that identity diversity only makes sense when there is satisfactory identity consistency.

\begin{table}[t]
\caption{Ablation study with Identity Consistency, Editability, Trusted Face Diversity and a proposed Identity Diversity. In addition, we also provide more parameter analysis in the supplementary material.}
\label{tab:ablation}
\setlength{\tabcolsep}{1.5pt}
\begin{tabular}{lccccc}
\toprule
            & Identity Cons. & Editability    & Trusted Div.               & Identity Div.            \\
\midrule            
Only $\mathcal{L}_{adv}$ & 0.078          & 0.299         & 0.013                   & \st{0.965}                    \\
Only $\mathcal{L}_{con}$ & 0.198          & 0.276         & 0.057                    & \st{0.741} \\
Ours        & \textbf{0.498}          & \textbf{0.332} & \textbf{0.140}  & \textbf{0.940}   \\
\bottomrule
\end{tabular}
\end{table}

As mentioned in Section~\ref{sec:3.2}, Only $\mathcal{L}_{adv}$ can generate prompt-aligned human images~(0.299 on Editability), but the generated faces from the same latent code $z$ are different~(0.078 on identity consistency). This is because learning the mapping $z\rightarrow v$ with only $\mathcal{L}_{adv}$ deceives the discriminator $D$, but still struggles to perceive contextual consistency. Only $\mathcal{L}_{con}$ is prone to mode collapse, producing similar identities for different $z$, which manifests as weaker identity diversity~(0.741). Notably, identity consistency is not significant under this setting. We attribute to the fact that direct L2 loss cannot reach the abstract objective~(i.e., identity consistency). When using $\mathcal{L}_{adv}$ and $\mathcal{L}_{con}$ together, IDE-GAN can generate diverse context-consistent pseudo identity embeddings, thereby achieving the best quantitative scores overall.

\subsection{Interpolation Property of IDE-GAN} The interpolation property of GANs is that interpolations between different randomly sampled latent codes in latent space can produce semantically smooth variations in image space~\cite{sainburg2018generative}. To evaluate whether our IDE-GAN carries this property, we randomly sample $z_1$ and $z_2$, and perform linear interpolation as shown in Figure~\ref{fig:interpolation}. 
IDE-GAN uses the interpolated latent codes to generate corresponding pseudo identity embeddings, respectively. Since the output space of IDE-GAN is embeddings instead of images, it cannot directly visualize the variations like traditional GANs~\cite{sainburg2018generative,karras2019style} in image space. So we insert these pseudo identity embeddings into Stable Diffusion to generate the corresponding images via the pipeline in Figure~\ref{fig:framework}(b). As shown in Figure~\ref{fig:interpolation}, CharacterFactory can produce continuous identity variations with the interpolations between different latent codes. And the interpolated latent codes~(e.g., $0.5z_1+0.5z_2$) can be chosen for further identity-consistent generation. It demonstrates that our IDE-GAN has satisfactory interpolation property and can be seamlessly integrated with Stable Diffusion.

\begin{figure}[t]
  \centering
  \includegraphics[width=1\linewidth]{"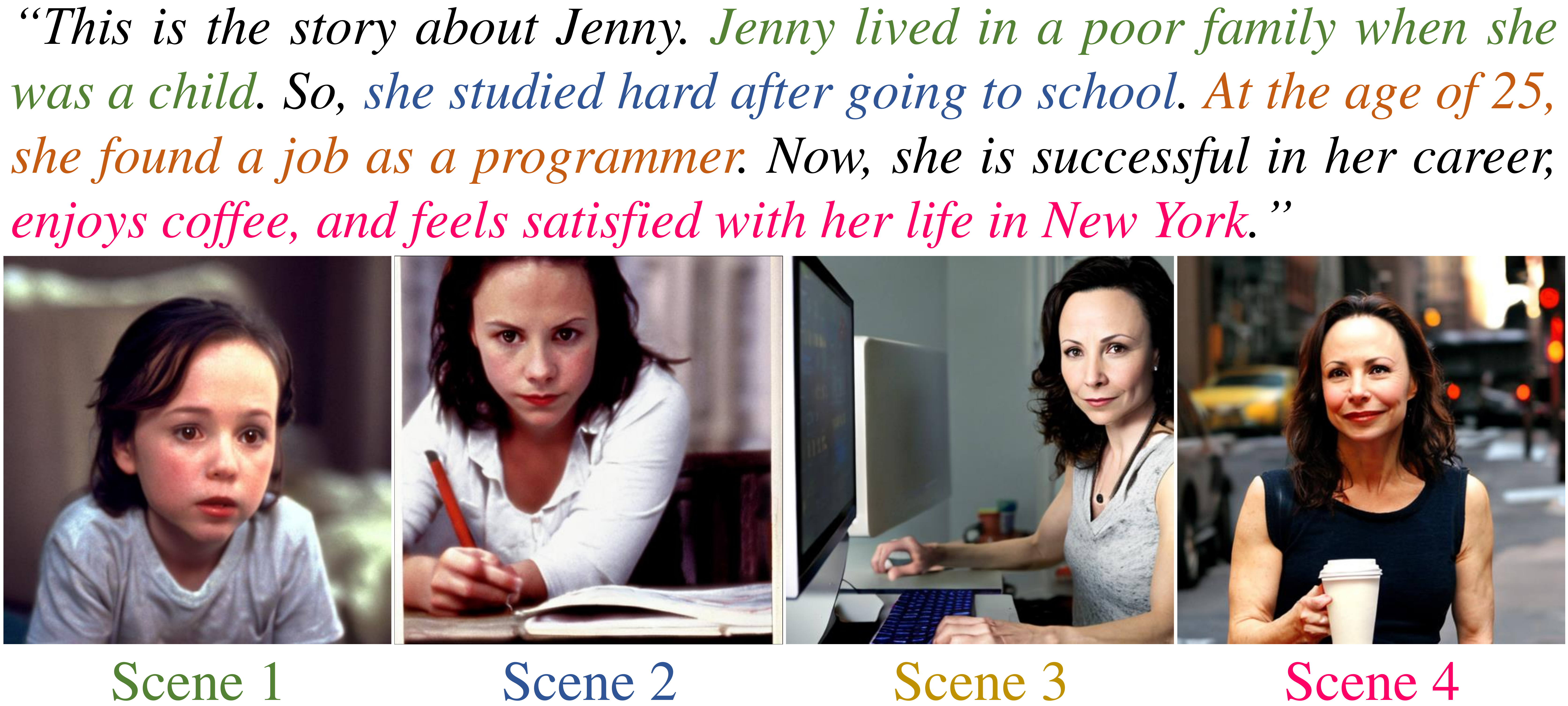"}
    \caption{Story Illustration. The proposed CharacterFactory can illustrate a story with the same character.}
  \label{fig:story}
\end{figure}

\subsection{Applications}
As shown in Figure~\ref{fig:teaser},~\ref{fig:story}, the proposed CharacterFactory can be used directly for various downstream tasks and is capable of broader extensions such as video/3D scenarios.

\noindent\textbf{Story Illustration.} In Figure~\ref{fig:story}, a full story can be divided into a set of text prompts for different scenes. CharacterFactory can create a new character to produce identity-consistent story illustrations.

\noindent\textbf{Stratified Sampling.} The proposed CharacterFactory can create diverse characters, such as different genders and races. Taking the gender as an example, we can categorize celeb names into ``Man'' and ``Woman'' to train Man-IDE-GAN and Woman-IDE-GAN separately, each of which can generate only the specified gender. Our generator $G$ is constructed with only two-layer MLPs, so that stratified sampling will not introduce excessive storage costs. More details can be found in the supplementary material.

\noindent\textbf{Virtual Humans in Image/Video/3D Generation.}
Currently, virtual human generation mainly includes 2D/3D facial reconstruction, talking-head generation and body/human movements~\cite{zhen2023human}, which typically rely on pre-existing images and lack scenario diversity and editability. And CharacterFactory can create new characters end-to-end and conduct identity-consistent virtual human image generation. In addition, since the pretrained Stable Diffusion 2.1 is fixed and the generated pseudo identity embeddings can be inserted into CLIP text transformer naturally, our method can collaborate with the SD-based plug-and-play modules. As shown in Figure~\ref{fig:teaser}, we integrate CharacterFactory with ControlNet-OpenPose~\cite{zhang2023adding, cao2017realtime}, ModelScopeT2V~\cite{wang2023modelscope} and LucidDreamer~\cite{liang2023luciddreamer} to implement identity-consistent virtual human image/video/3D generation.

\noindent\textbf{Identity-Consistent Dateset Construction.} 
Some human-centric subject-driven generation methods~\cite{li2023photomaker,chen2023dreamidentity} construct large-scale celeb datasets for training. PhotoMaker~\cite{li2023photomaker} crawls celeb 
photos from the Internet and DreamIdentity~\cite{chen2023dreamidentity} uses text prompts containing celeb names to drive Stable Diffusion to generate celeb images. Their constructed data includes only celebs, leading to a limited number of identities. Notably, the proposed CharacterFactory can use diverse text prompts to generate identity-consistent images infinitely for dataset construction. Furthermore, collaboration with the mentioned SD-based plug-and-play modules can construct identity-consistent video/3D datasets.

\section{Conclusion}
In this work, we propose CharacterFactory, to unlock the end-to-end identity-consistent generation ability for diffusion models. It consists of an Identity-Embedding GAN~(IDE-GAN) for learning the mapping from a latent space to the celeb embedding space and a context-consistent loss for identity consistency. It takes only 10 minutes for training and 3 seconds for end-to-end inference. Extensive quantitative and qualitative experiments demonstrate the superiority of CharacterFactory. Besides, we also present that our method can empower many interesting applications.

\input{main.bbl}


\clearpage
\appendix

\twocolumn[{
\renewcommand\twocolumn[1][]{#1}
\begin{center}
\centering
    \captionsetup{type=figure}
    \includegraphics[width=1\linewidth]{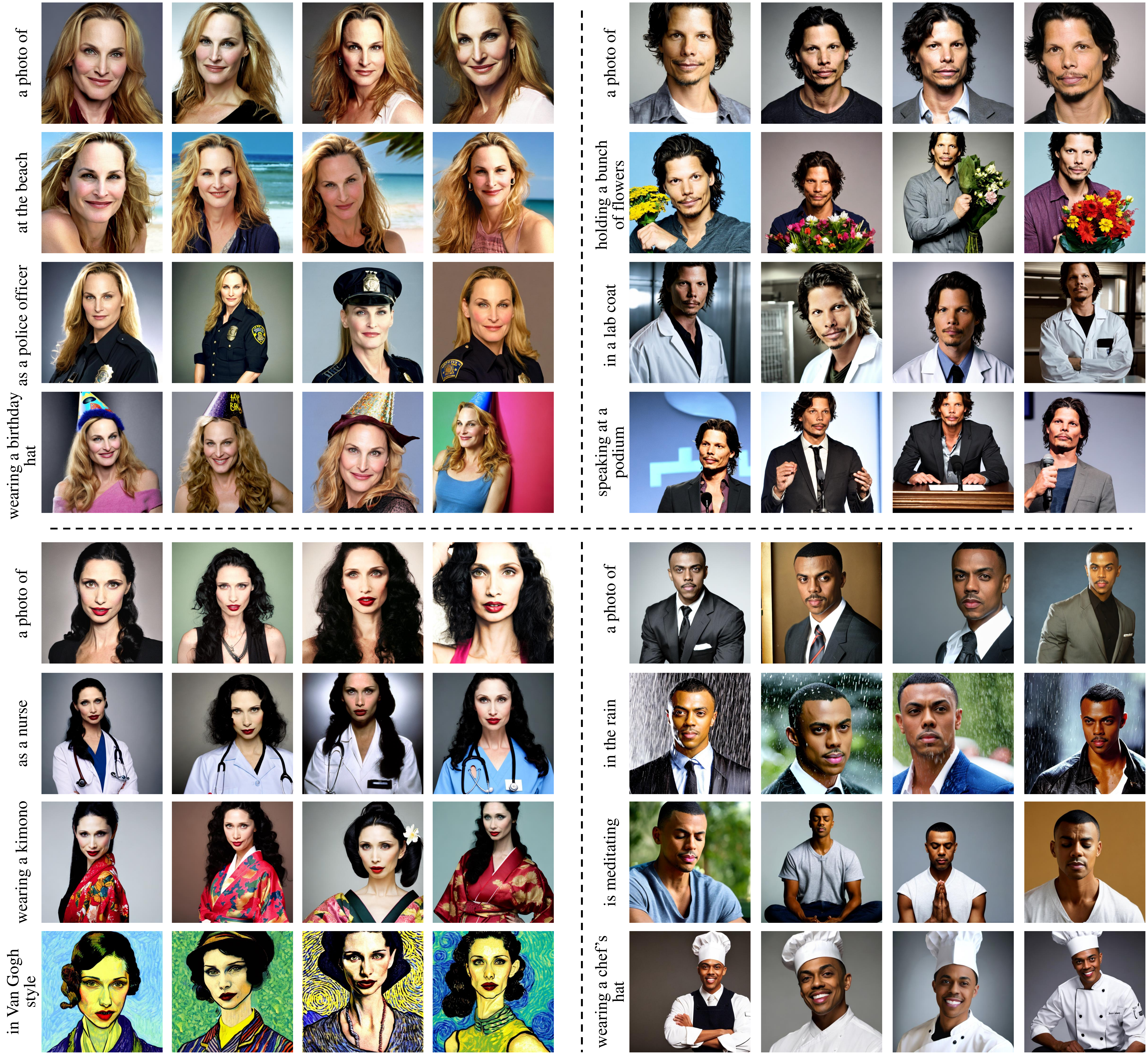}
    \captionof{figure}{More identity-consistent character generation results by the proposed CharacterFactory. The placeholders $s_1^*$, $s_2^*$ of prompts such as ``a photo of $s_1^*$ $s_2^*$'' are omitted in this work for brevity~(\textit{Zoom in for the best view}).
    }
    \label{fig:supp_teaser}
\end{center}
}]



\begin{figure*}[!t]
  \centering
  \includegraphics[width=1\linewidth]{"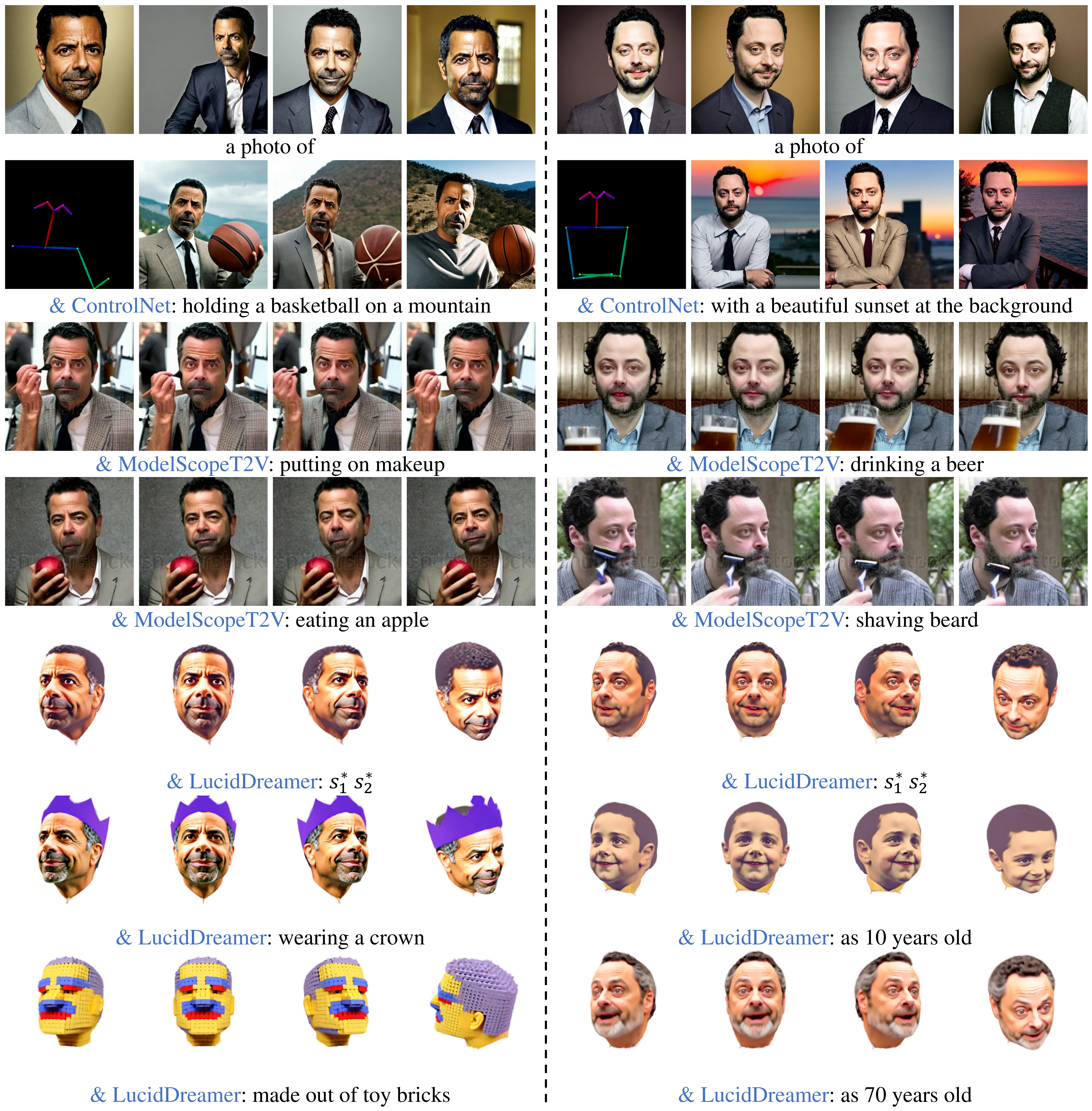"}
      \caption{More identity-consistent Image/Video/3D generation results by the proposed CharacterFactory with ControlNet~\cite{zhang2023adding}, ModelScopeT2V~\cite{wang2023modelscope} and LucidDreamer~\cite{liang2023luciddreamer}~(\textit{Zoom in for the best view}).}
  \label{fig:supp_video_3d}
\end{figure*}

\section{More Results}
\begin{figure}[!t]
  \centering
  \includegraphics[width=1\linewidth]{"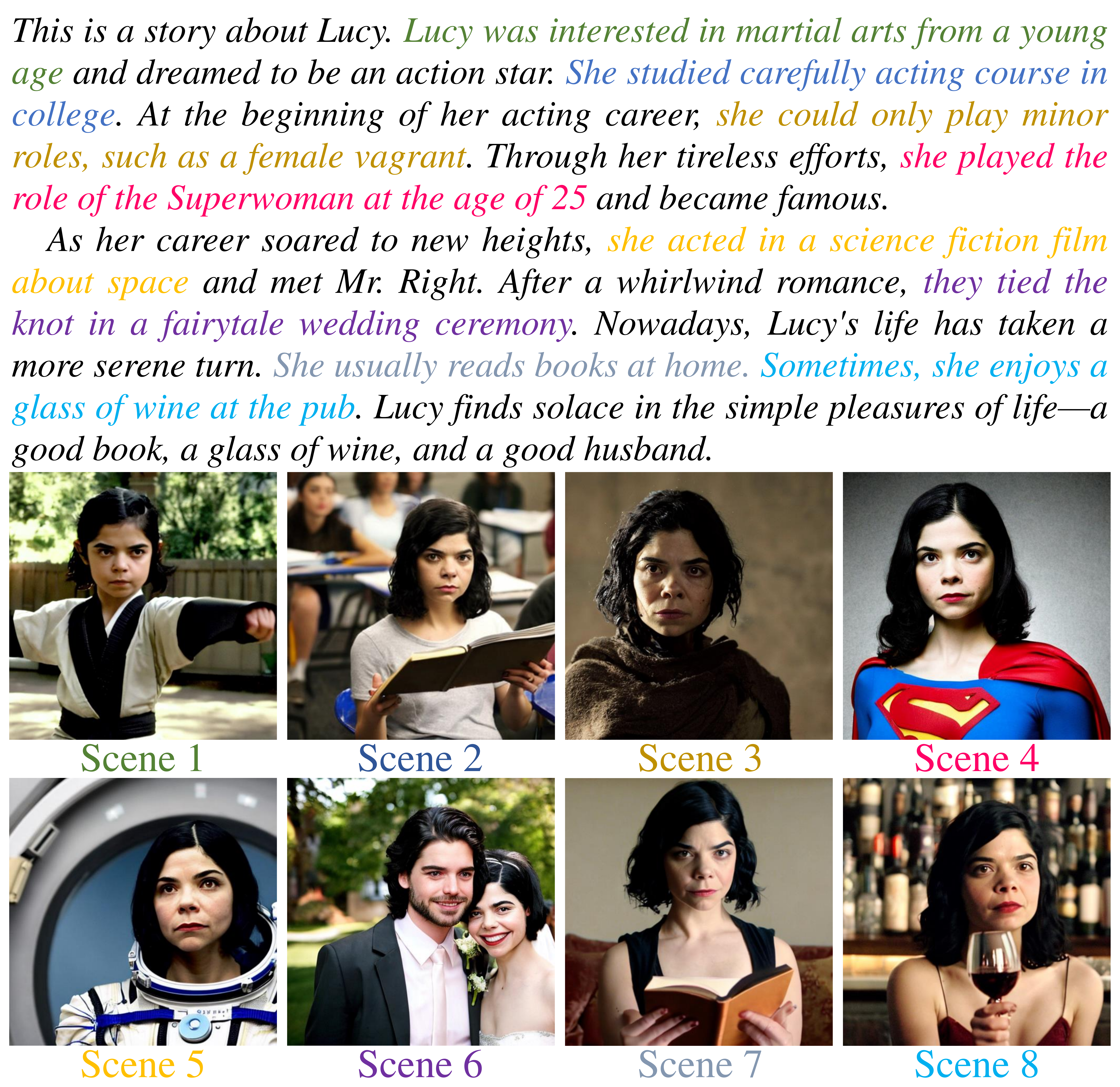"}
      \caption{We provide more identity-consistent illustrations for a longer continuous story to further demonstrate the availability of this application.}
  \label{fig:long_story}
\end{figure}

\noindent\textbf{More Consistent Image/Video/3D Generation Results.} As shown in Figure~\ref{fig:supp_teaser}, we show more identity-consistent character generation results in diverse text prompts covering actions, backgrounds, decorations and styles. The stunning generation results demonstrate the powerful ability of the proposed CharacterFactory to create new characters. We also show more illustrations for a longer story as shown in Figure~\ref{fig:long_story}.

In addition, we provide identity-consistent Image/Video/3D generation results as shown in Figure~\ref{fig:supp_video_3d}. Seamless integration with ControlNet~\cite{zhang2023adding}, ModelScopeT2V~\cite{wang2023modelscope} and LucidDreamer~\cite{liang2023luciddreamer} empowers identity-consistent virtual human image/video/3D generation.

\section{More Implementation Details}
\noindent\textbf{The Dimension of Latent Code $z$.} 
Similar to traditional GAN models~\cite{karras2019style,fu2022stylegan}, the input of our Identity-Embedding GAN~(IDE-GAN) is randomly sampled from a latent space, i.e., $z\in\mathcal{N}(0,\mathbf{I})$. Differently, our objective is to learn a mapping from the latent space to the celeb embedding space, i.e., $G: z\rightarrow v$. The CLIP word embedding's dimension of Stable Diffusion 2.1 is 1024, so the dimension of IDE-GAN's output $[v_1^*, v_2^*]$ is $[2, 1024]$. Based on this, we conduct experiments for the dimension of $z$, as shown in Table~\ref{tab:z}. Overall, IDE-GAN can generate consistent pseudo identity conforming to the given text prompts with different dimensions of $z$. However, we empirically find that a larger dimension of $z$ introduces more trainable parameters, leading to harder convergence and degraded identity consistency. A smaller $z$ represents narrower input space, resulting in poor identity diversity~(this metric is designed in Section~4.4 for the diversity of generated identities). Therefore, we choose the setting of ``Dim(z)=64'' to trade off identity consistency and identity diversity, which performs well across the board.

\begin{table}[t]
\caption{Parameter analysis for the dimension of latent code $z$ on Editability, Identity Consistency~(i.e., ID Cons.), Identity Diversity~(i.e., ID Div.) and Image Quality. }
\label{tab:z}
\setlength{\tabcolsep}{4pt}
\begin{tabular}{lcccc}
\toprule
           & Editability$\uparrow$    & ID Cons.$\uparrow$  & ID Div.$\uparrow$  & Img Quality$\downarrow$  \\
\midrule      
Dim(z)=8  & 0.317          & 0.496          & 0.902          & 23.61          \\
Dim(z)=16  & 0.322          & 0.493          & 0.926          & 23.10          \\
Dim(z)=32  & 0.328          & 0.492          & 0.931          & 22.77          \\
\textbf{Dim(z)=64}  & \textbf{0.332} & \textbf{0.498}  & 0.940          & \textbf{22.58}          \\
Dim(z)=128 & 0.323          & 0.475 & 0.939 & 22.66 \\
Dim(z)=256 & 0.318          & 0.462          & 0.955 & 22.92        \\
Dim(z)=512 & 0.320          & 0.454          & \textbf{0.962} & 23.52        \\
\bottomrule
\end{tabular}
\end{table}

\noindent\textbf{More Details of IDE-GAN.} IDE-GAN is made up of MLPs, where the generator is $64-2048-2048$ MLP architecture and the discriminator is $2048-512-256-1$ MLP architecture. In addition to the ablation experiments in Section~4.4 of the main paper, we further explore the effect of AdaIN~\cite{wang2024stableidentity}. Due to the instability of GAN training, the generated embeddings have an unrestricted value range and fail to naturally insert into diffusion models under the setting of ``w/o AdaIN''~(decent results), as shown in Table~\ref{tab:wo_adain}. Since each dimension of word embedding has a different distribution, the AdaIN is introduced to contribute to the distribution alignment. Specifically, this operation helps the generated pseudo identity embeddings land more naturally into the celeb embedding space, so that the generator can deceive the discriminator more easily. Therefore, the introduced AdaIN will further unleash the prompt-aligned identity-consistent generation ability of IDE-GAN.

\begin{table}[t]
\caption{Ablation study for the introduced AdaIN. The setting of ``w/o AdaIN'' means using $\mathcal{L}_{adv} + \mathcal{L}_{con}$ without AdaIN.}
\label{tab:wo_adain}
\begin{tabular}{lccc}
\toprule
          & ID Cons.$\uparrow$ & Editability$\uparrow$    & Img Quality$\downarrow$  \\
\midrule          
w/o AdaIN & 0.187          & 0.298         & 33.67          \\
Ours      & \textbf{0.498} & \textbf{0.332} & \textbf{22.58} \\
\bottomrule
\end{tabular}
\end{table}

\noindent\textbf{More Details of Stratified Sampling.} Since celeb names can be categorized into different groups based on gender, race or other characteristics, IDE-GAN can be trained on the celeb embeddings of different groups to achieve stratified sampling. Taking the gender as an example, we manually categorize 326 celeb names into 226 men and 100 women to train Man-IDE-GAN and Woman-IDE-GAN for creating male characters and female characters separately. As shown in Table~\ref{tab:man_woman}, we provide the quantitative results of Man-IDE-GAN and Woman-IDE-GAN. It can be observed that the performance of stratified sampling is satisfactory. On the other hand, when the number of celeb names decreases, Man-IDE-GAN and Woman-IDE-GAN show acceptable performance compared to the original IDE-GAN and take fewer training steps.

\begin{table}[t]
\caption{Quantitative results of Man-IDE-GAN and Woman-IDE-GAN for stratified sampling on gender. We also provide the training steps of each setting.}
\setlength{\tabcolsep}{1pt}
\label{tab:man_woman}
\begin{tabular}{lccccc}
\toprule
              & Editability & ID Cons. & ID Div. & Img Quality  & Steps\\
\midrule               
Man-IDE-GAN   & 0.320       & 0.477    & 0.934   & 25.36  & 7000     \\
Woman-IDE-GAN & 0.318       & 0.451    & 0.930   & 27.41  & 6000     \\
IDE-GAN       & 0.332       & 0.498    & 0.962   & 22.58  & 10000    \\
\bottomrule
\end{tabular}
\end{table}

\section{Limitations}
Although the proposed CharacterFactory enables end-to-end identity-consistent character generation, it is not free of limitations. First, GAN-based methods can generate high-quality images, but these images may look unnatural with artifacts~\cite{karras2019style,fu2022stylegan}. Similarly, our IDE-GAN may produce suboptimal embeddings, lacking consistency. Second, since the UNet of Stable Diffusion~(SD) is fixed, we inherit not only the powerful generation ability, but also its drawbacks, such as hand anomalies~\cite{zhang2023detecting}.

\end{document}